%% file: root.tex
\documentclass[letterpaper, 10 pt, conference]{ieeeconf}

\IEEEoverridecommandlockouts

\input{packages}

\newcommand \colorindicator[1]{%
	{\textcolor{#1}{$\blacksquare\!\!\!\!\blacksquare$}}%
}

\title{\LARGE \bf
An Instance Segmentation Dataset of Yeast Cells in Microstructures
}

\author{Christoph Reich\textsuperscript{1,}*, Tim Prangemeier\textsuperscript{1,}*, Andr{\'e} O. Fran{\c{c}}ani\textsuperscript{2}, Heinz Koeppl\textsuperscript{1,\ddag}%
\thanks{* Christoph Reich \& Tim Prangemeier - both authors contributed equally}%
\thanks{\textsuperscript{\ddag} Correspondence: \href{mailto:heinz.koeppl@tu-darmstadt.de}{\tt\small heinz.koeppl@tu-darmstadt.de}}%
\thanks{\textsuperscript{1} Centre for Synthetic Biology, Department of Electrical Engineering and Information Technology, Technische Universit\"at Darmstadt}%
\thanks{\textsuperscript{2} Aeronautics Institute of Technology, work done while at TU Darmstadt}%
}

\begin{document}

\maketitle
\thispagestyle{empty}
\pagestyle{empty}

\begin{abstract}

Extracting single-cell information from microscopy data requires accurate instance-wise segmentations. Obtaining pixel-wise segmentations from microscopy imagery remains a challenging task, especially with the added complexity of microstructured environments. This paper presents a novel dataset for segmenting yeast cells in microstructures. We offer pixel-wise instance segmentation labels for both cells and trap microstructures. In total, we release 493 densely annotated microscopy images. To facilitate a unified comparison between novel segmentation algorithms, we propose a standardized evaluation strategy for our dataset. The aim of the dataset and evaluation strategy is to facilitate the development of new cell segmentation approaches. The dataset is publicly available at \url{https://christophreich1996.github.io/yeast_in_microstructures_dataset/}.

\end{abstract}


\input{content/introduction}
\input{content/dataset}
\input{content/performance_evaluation}

\input{content/conclusion_outlook}

\section*{Acknowledgment}
We thank Christoph Hoog Antink for insightful discussions, Klaus-Dieter Voss for aid with the microfluidics fabrication, and Jan Basrawi for contributing to data labelling. 

This work was supported by the Landesoffensive f{\"u}r wissenschaftliche Exzellenz as part of the LOEWE Schwerpunkt CompuGene. H.K. acknowledges the support from the European Research Council (ERC) with the consolidator grant CONSYN (nr. 773196). C.R. acknowledges the support of NEC Laboratories America, Inc.


\bibliographystyle{IEEEtran}
\bibliography{references.bib}

\end{document}

%% file: packages.tex
\usepackage{graphicx}
\usepackage{amsmath}
\usepackage{amssymb}
\let\labelindent\relax
\usepackage[inline]{enumitem}
\usepackage{booktabs}
\usepackage{tabularx}
\usepackage{multirow}
\usepackage{makecell}
\usepackage[export]{adjustbox}
\usepackage[per-mode=symbol]{siunitx}
\usepackage{float}
\usepackage[noadjust]{cite}
\usepackage{pgfplots}
\usepackage{xcolor}
\usepackage{tikz}
\usepackage{pgfplotstable}
\usepackage[hidelinks, pdfborder = {0 0 0}]{hyperref}
\usepackage[capitalise]{cleveref}

\usetikzlibrary{pgfplots.groupplots}
\usepgfplotslibrary{statistics}
\usepgfplotslibrary{groupplots,dateplot}
\pgfplotsset{compat=1.3}

\definecolor{tud0d}{cmyk/RGB/HTML}{0,0,0,.8/83,83,83/535353}
\definecolor{tud0c}{cmyk/RGB/HTML}{0,0,0,.6/137,137,137/898989}
\definecolor{tud0b}{cmyk/RGB/HTML}{0,0,0,.4/181,181,181/B5B5B5}
\definecolor{tud0a}{cmyk/RGB/HTML}{0,0,0,.2/220,220,220/DCDCDC}
\definecolor{tud1a}{cmyk/RGB/HTML}{.7,.4,0,0/93,133,195/5D85C3}
\definecolor{tud2a}{cmyk/RGB/HTML}{0.8,.2,0,0/0,156,218/009CDA}
\definecolor{tud3a}{cmyk/RGB/HTML}{0.7,0,.5,0/80,182,149/50B695}
\definecolor{tud4a}{cmyk/RGB/HTML}{.4,0,.8,0/175,204,80/AFCC50}
\definecolor{tud5a}{cmyk/RGB/HTML}{.2,0,.8,0/221,223,72/DDDF48}
\definecolor{tud6a}{cmyk/RGB/HTML}{0,.1,.7,0/255,224,92/FFE05C}
\definecolor{tud7a}{cmyk/RGB/HTML}{0,.3,.8,0/248,186,60/F8BA3C}
\definecolor{tud8a}{cmyk/RGB/HTML}{0,.6,.8,0 /238,122,52/EE7A34}
\definecolor{tud9a}{cmyk/RGB/HTML}{0,.8,.7,0/233,80,62/E9503E}
\definecolor{tud10a}{cmyk/RGB/HTML}{.2,.9,0,0/201,48,142/C9308E}
\definecolor{tud11a}{cmyk/RGB/HTML}{.6,.8,0,0/128,69,151/804597}
\definecolor{tud1b}{cmyk/RGB/HTML}{1,.6,0,0/0,90,169/005AA9}
\definecolor{tud2b}{cmyk/RGB/HTML}{1,.3,0,0/0,131,204/0083CC}
\definecolor{tud3b}{cmyk/RGB/HTML}{1,0,.6,0/0,157,129/009D81}
\definecolor{tud4b}{cmyk/RGB/HTML}{.5,0,1,0/153,192,0/99C000}
\definecolor{tud5b}{cmyk/RGB/HTML}{.3,0,1,0/201,212,0/C9D400}
\definecolor{tud6b}{cmyk/RGB/HTML}{0,.2,1,0/253,202,0/FDCA00}
\definecolor{tud7b}{cmyk/RGB/HTML}{0,.4,1,0/245,163,0/F5A300}
\definecolor{tud8b}{cmyk/RGB/HTML}{0,.7,1,0/236,101,0/EC6500}
\definecolor{tud9b}{cmyk/RGB/HTML}{0,1,.9,0/230,0,26/E6001A}
\definecolor{tud10b}{cmyk/RGB/HTML}{.4,1,0,0/166,0,132/A60084}
\definecolor{tud11b}{cmyk/RGB/HTML}{.7,1,0,0/114,16,133/721085}
\definecolor{tud1c}{cmyk/RGB/HTML}{1,.7,.2,0/0,78,138/004E8A}
\definecolor{tud2c}{cmyk/RGB/HTML}{1,.5,.2,0/0,104,157/00689D}
\definecolor{tud3c}{cmyk/RGB/HTML}{1,.2,.6,0/0,136,119/008877}
\definecolor{tud4c}{cmyk/RGB/HTML}{.6,.1,1,0/127,171,22/7FAB16}
\definecolor{tud5c}{cmyk/RGB/HTML}{.4,.1,1,0/177,189,0/B1BD00}
\definecolor{tud6c}{cmyk/RGB/HTML}{.2,.3,1,0/215,172,0/D7AC00}
\definecolor{tud7c}{cmyk/RGB/HTML}{.2,.5,1,0/210,135,0/D28700}
\definecolor{tud8c}{cmyk/RGB/HTML}{.2,.8,1,0/204,76,3/CC4C03}
\definecolor{tud9c}{cmyk/RGB/HTML}{.3,1,.9,0/185,15,34/B90F22}
\definecolor{tud10c}{cmyk/RGB/HTML}{.5,1,.3,0/149,17,105/951169}
\definecolor{tud11c}{cmyk/RGB/HTML}{.8,1,.2,0/97,28,115/611C73}
\definecolor{tud1d}{cmyk/RGB/HTML}{1,.9,.3,0/36,53,114/243572}
\definecolor{tud2d}{cmyk/RGB/HTML}{1,.7,.4,0/0,78,115/004E73}
\definecolor{tud3d}{cmyk/RGB/HTML}{1,.4,.7,0/0,113,94/00715E}
\definecolor{tud4d}{cmyk/RGB/HTML}{.7,.3,1,0/106,139,55/6A8B22}
\definecolor{tud5d}{cmyk/RGB/HTML}{.5,.2,1,0/153,166,4/99A604}
\definecolor{tud6d}{cmyk/RGB/HTML}{.4,.4,1,0/174,142,0/AE8E00}
\definecolor{tud7d}{cmyk/RGB/HTML}{.3,.6,1,0/190,111,0/BE6F00}
\definecolor{tud8d}{cmyk/RGB/HTML}{.4,.8,1,0/169,73,19/A94913}
\definecolor{tud9d}{cmyk/RGB/HTML}{.5,1,.9,0/156,28,38/961C26}
\definecolor{tud10d}{cmyk/RGB/HTML}{.7,1,.5,0/115,32,84/732054}
\definecolor{tud11d}{cmyk/RGB/HTML}{.9,1,.3,0/76,34,106/4C226A}

\definecolor{cell}{rgb}{1., 0.5, 0.90980392}
\definecolor{trap}{rgb}{0.25, 0.25, 0.25}

%% file: content/introduction.tex
\section{Introduction} \label{sec:introduction}

\begin{figure}[th!]
    \centering
    \includegraphics[width=0.25\columnwidth]{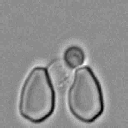}\includegraphics[width=0.25\columnwidth]{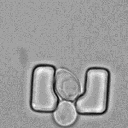}\includegraphics[width=0.25\columnwidth]{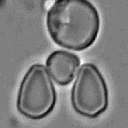}\includegraphics[width=0.25\columnwidth]{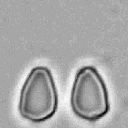}\\%
    \includegraphics[width=0.25\columnwidth]{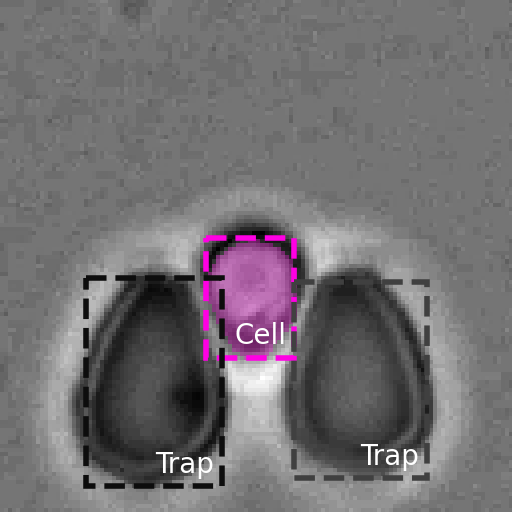}\includegraphics[width=0.25\columnwidth]{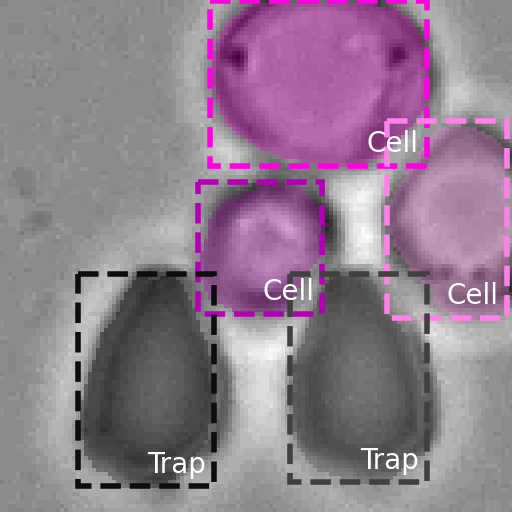}\includegraphics[width=0.25\columnwidth]{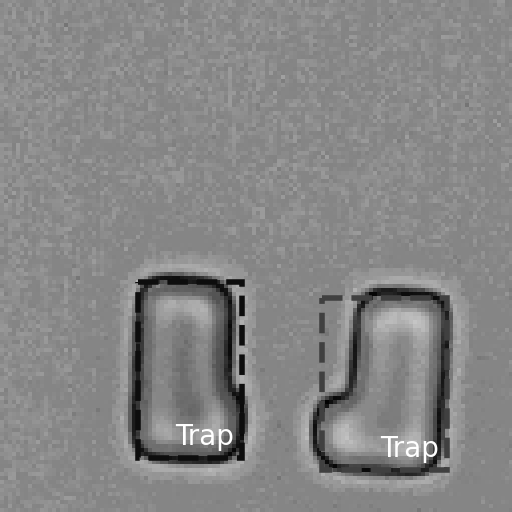}\includegraphics[width=0.25\columnwidth]{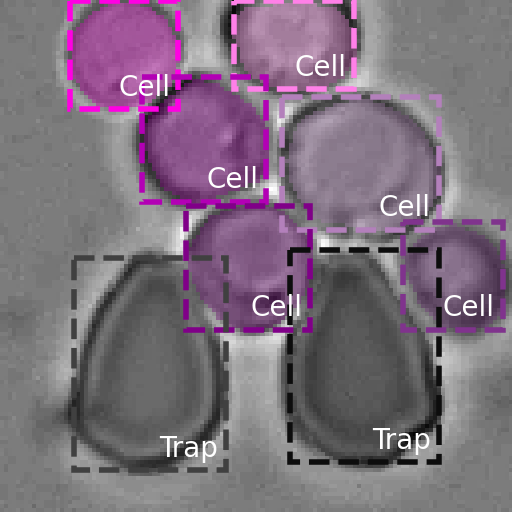}\\%
    \includegraphics[width=0.25\columnwidth]{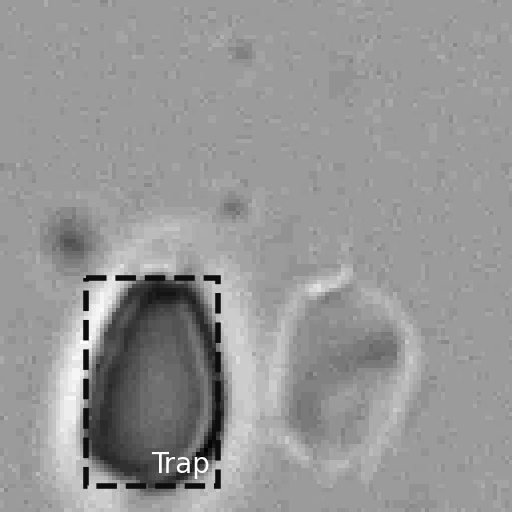}\includegraphics[width=0.25\columnwidth]{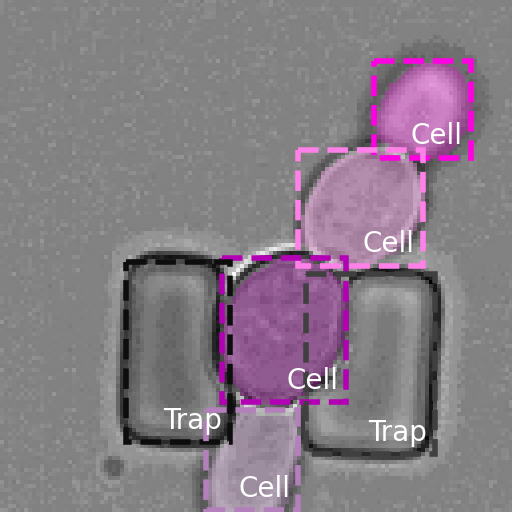}\includegraphics[width=0.25\columnwidth]{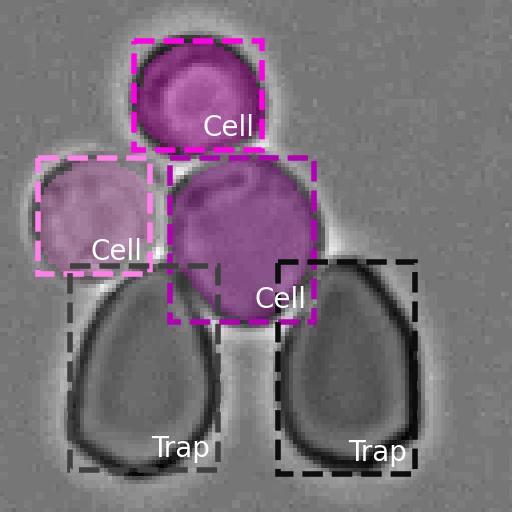}\includegraphics[width=0.25\columnwidth]{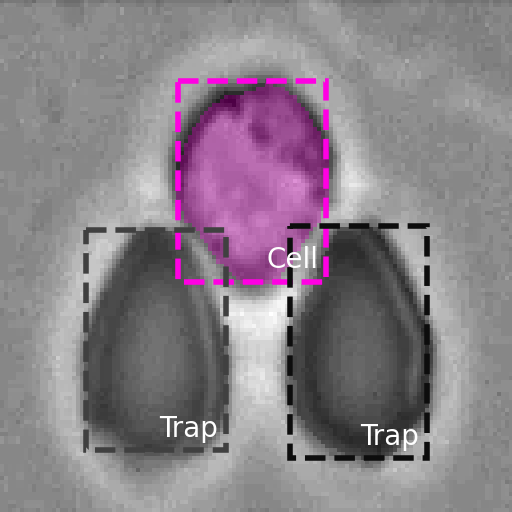}%
    \caption{Samples of our yest cells in microstructures dataset. The top row show unlabelled brightfield imagery to demonstrate the visual similarity between the cells and similarly sized microsctuctures. In the following rows, the instance segmentation labels are overlayed onto the the brightfield microscopy imagery (grayscale). A bounding box and object class label denotes each object for clarity. Shades of pink (\colorindicator{cell}) indicate individual cell instances and shades of (dark) gray (\colorindicator{gray}) indicate microstructures (traps).}
    \label{fig:firstfig}
    \vspace{-0.5em}
\end{figure}

Many biomedical applications require the detection and segmentation of individual cells in microscopy imagery~\cite{Meijering2012}. For example, analyzing the cellular processes of living cells in time-lapse fluorescence microscopy (TFLM) experiments, requires accurate pixel-level segmentations of individual cells~\cite{Pepperkok2006, Bakker2018, Prangemeier2020, Aspert2022}. Most applications require each cell to be segmented and identified as an unique entity or instance~\cite{Prangemeier2020b}. Instance segmentation is the task of detecting, segmenting, and classifying each object instance in an image~\cite{Cordts2016}. While powerful cell segmentation algorithms have been proposed recently (\textit{e.g.}~\cite{Schmidt2018, Stringer2021}), segmenting cells in microstructured environments remains challenging, due to the perceptual similarity of microstructurs and cells (\textit{cf.} \cref{fig:firstfig}) \cite{Bakker2018, Francani2022, Prangemeier2022}.\looseness=-1

The vast majority of current state-of-the-art segmentation algorithms utilize deep neural networks~\cite{He2017, Antink2020, Reich2021b, Stringer2021}. A key factor driving the development of deep learning-based segmentation algorithms is the widespread availability of pixel-wise annotated datasets. Examples include Microsoft COCO~\cite{Lin2014}, Cityscapes~\cite{Cordts2016}, ADE20K~\cite{Zhou2017}, and the 2018 Data Science Bowl dataset~\cite{Caicedo2019}. While general cell segmentation datasets are available (\textit{e.g.}~\cite{Caicedo2019, Parekh2019}), we are not aware of any instance segmentation dataset of cells in microstructures with dense annotations.

In this paper, we present and publicly release a dataset of yeast (\textit{Saccharomyces cerevisiae}) cells in microstructures with instance segmentation annotations. The dataset is comprised of 493 densely annotated brightfield microscopy images of different TLFM experiments (\textit{cf.} \cref{fig:firstfig}). To facilitate a fair comparison between novel segmentation approaches we also propose a standardized performance evaluation strategy. The PyTorch~\cite{Paszke2019} code for performance evaluation is publicly available at \url{https://github.com/ChristophReich1996/Yeast-in-Microstructures-Dataset}.

%% file: content/dataset.tex
\section{Dataset} \label{sec:dataset}

When developing a segmentation dataset, numerous design choices have to be made. This section will give a detailed overview of these design decisions. First, we describe the data acquisition and annotation process. Second, we will introduce the core features and statistics of our dataset. Finally, we describe how our dataset is split for training, validation, and testing.

\subsection{Data Acquisition}

\begin{figure}[th!]
    \centering
    \includegraphics[width=0.745\columnwidth]{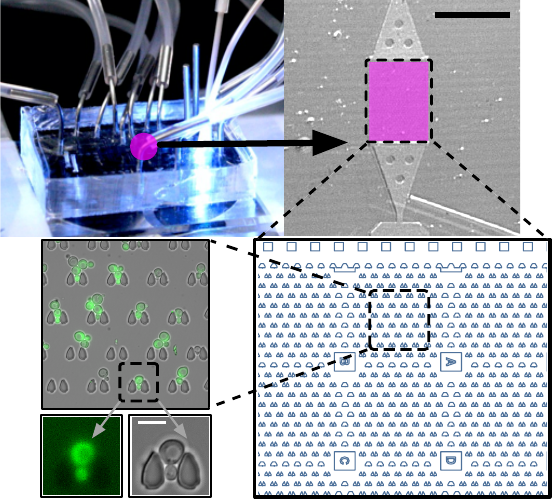}
    \caption{TLFM experiment setup for single-cell fluorescence measurement. A microfluidic chip sits atop the microscope table (top left). The trap chamber (pink \colorindicator{cell} on the top right) contains approximately one thousand traps. We extract cropped specimen images from the fluorescence and brightfield channels (bottom left), that include a pair of trap microstructures and cells. The brightfield channel is used for segmentation (and in this dataset). The black scale bar is $1\si{\milli\meter}$, the white scale bar is $10\si{\micro\meter}$.}
    \label{fig:complexchip}
    \vspace{-0.5em}
\end{figure}

We chose two common yeast trap microstructure geometries and drew data from a wide range of experiments performed in our lab to generate our dataset (\textit{cf.} \cref{fig:firstfig}) \cite{Crane2014, Reich2021a}. An overview of the experimental setup atop the microscope table is given in \cref{fig:complexchip}. We designed and fabricated trap microchips for long-term cultures of yeast cells (\textit{cf.} \cref{fig:complexchip}). We recorded brightfield microscopy images of the living yeast cells confined to the microfluidic chip over many hours. A constant flow of yeast growth media hydrodynamically traps the cells in the microstructure pairs~\cite{Prangemeier2018, Prangemeier2020}.

We extracted $493$ specimen images, each centered on a single trap microstructure pair with a resolution of $128\times 128$, from the higher resolution raw data, as is common practice (\textit{cf.} \cref{fig:complexchip})~\cite{Bakker2018, Reich2021a}. We differentiate between two subsets, one for each of the trap geometries employed (roughly oval shaped \textit{regular traps}, and the \textit{L traps}). In order to increase the robustness and range of applicability of the models trained on this data, we include variations in trap type, debris, focal shift illumination levels, and yeast morphology. Our dataset captures the most common yeast-trap configurations: \begin{enumerate*}[label=(\roman*), font=\itshape]
\item empty traps
\item single cells (with daughter) and
\item multiple cells~\cite{Prangemeier2020b}.
\end{enumerate*} Our dataset also includes edge cases such as broken traps (\textit{cf.} \cref{fig:firstfig} bottom left).

\subsection{Data Annotation}
We present $493$ pixel-level annotated images, with pixel-wise instance-level annotations of both cells and microstructures (traps). Each pixel, not labeled as a cell or tap, is considered as background. Note, since our labels only include a single background class, our instance segmentation labels can also be seen as panoptic labels~\cite{Kirillov2019}. This property is later used for performance evaluation (\textit{cf.} \cref{sec:performanceevaluation}).

\begin{figure}[th!]
    \centering
    \frame{\frame{\includegraphics[width=0.162\columnwidth]{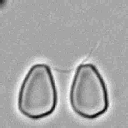}\includegraphics[width=0.162\columnwidth]{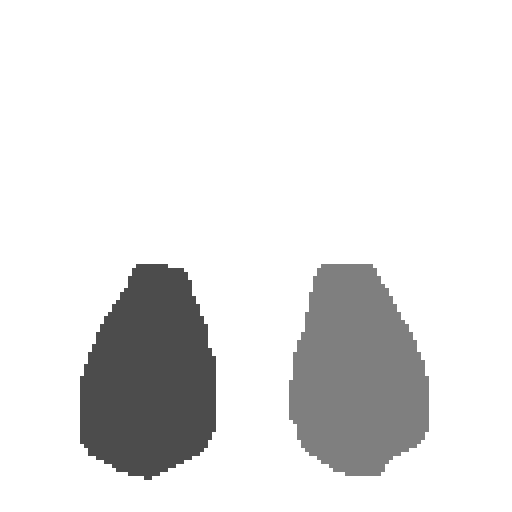}\includegraphics[width=0.162\columnwidth]{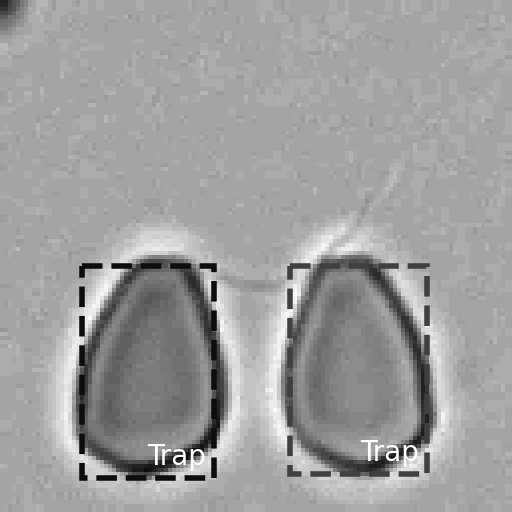}}}\hfill\frame{\frame{\includegraphics[width=0.162\columnwidth]{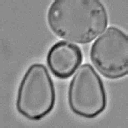}\includegraphics[width=0.162\columnwidth]{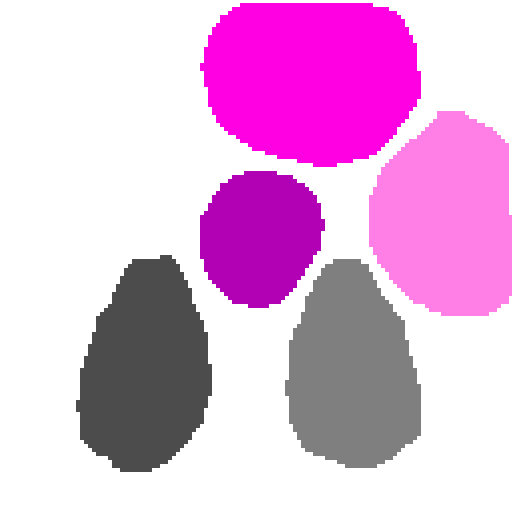}\includegraphics[width=0.162\columnwidth]{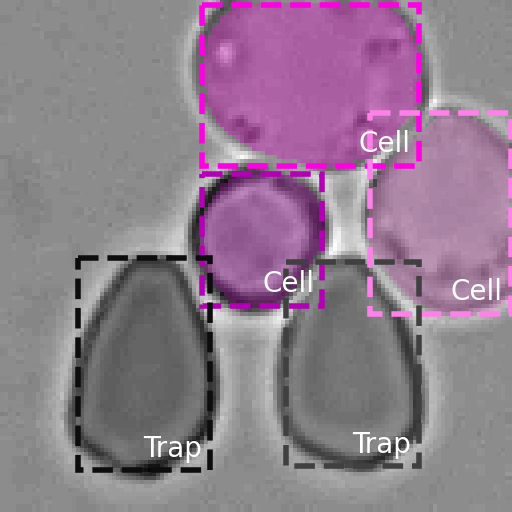}}}\\[0.19cm]
    \frame{\frame{\includegraphics[width=0.162\columnwidth]{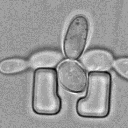}\includegraphics[width=0.162\columnwidth]{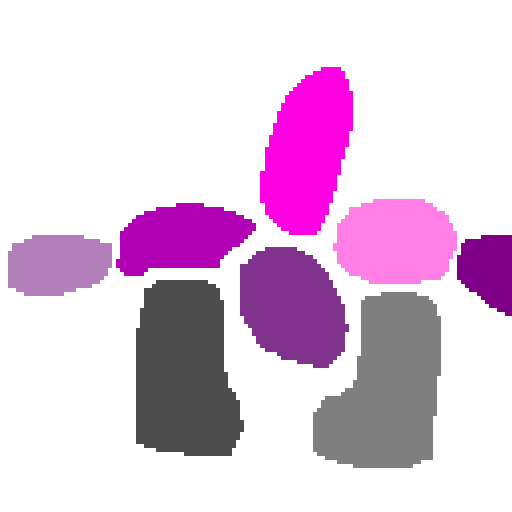}\includegraphics[width=0.162\columnwidth]{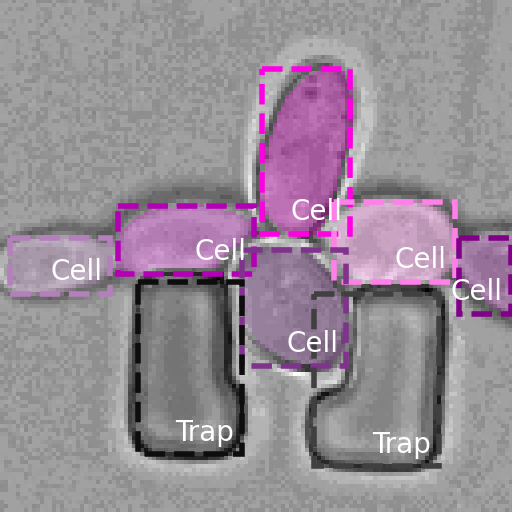}}}\hfill\frame{\frame{\includegraphics[width=0.162\columnwidth]{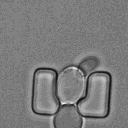}\includegraphics[width=0.162\columnwidth]{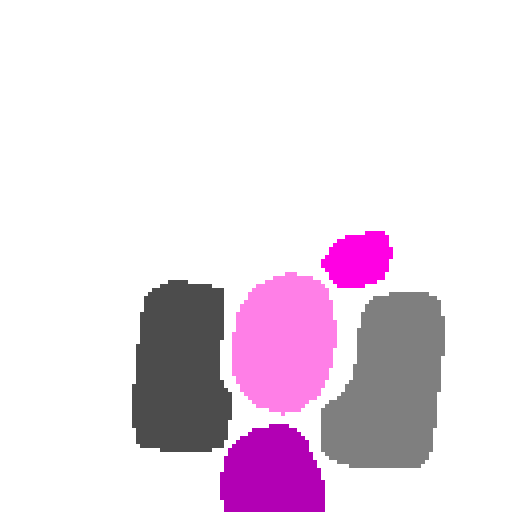}\includegraphics[width=0.162\columnwidth]{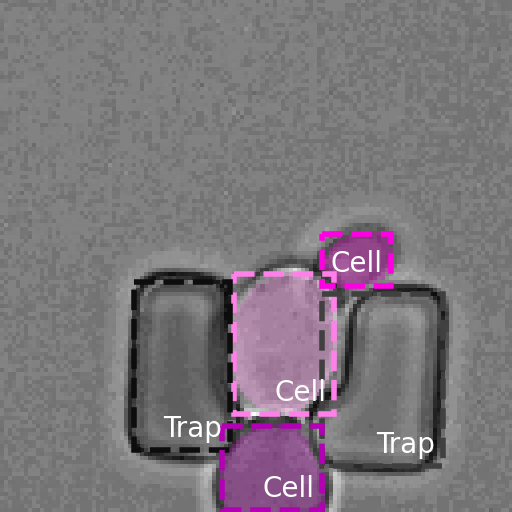}}}
    \caption{Brightfield microscopy imagery of yeast cells and microstructures with the corresponding labels. Brightfield images on the left, instance segmentation label in the middle, and an overlay of the brightfield images and labels on the right. Shades of gray (\colorindicator{gray}) indicate different instances of microstructures (trap). Cell instances are visualized in shades of pink (\colorindicator{cell}). The background is white.}
    \label{fig:labels}
    \vspace{-0.5em}
\end{figure}

All annotations were acquired manually. For every object instance (cells and traps), we annotate the object class and the pixels belonging to the object. Note that we assume no overlapping cell and trap instances (seeing as the chips are designed to prevent any overlap). This annotation process results in a binary segmentation map and classification for each object instance. \cref{fig:labels} showcases both brightfield images and our manual instance-wise annotations.

In most cases, cell instances can easily be distinguished during labeling. In the case of budding, in which a daughter cell pinches off a mother cell, labeling the growing daughter cell is non-trivial~\cite{Duina2014}. We decided to annotate the daughter cell as a separate instance if it is clearly separated from the mother cell. An example of this is given in \cref{fig:labels} (bottom right). The detection of daughter cells, can, for example, aid in determining the cell fitness.

While most applications, such as analyzing the cellular process of living cells, mainly require the segmentation of cells, we decided to also include annotations of microstructures in our dataset. The reason for this decision is twofold. First, when knowing both the position cells and traps it can be determined which cells are hydrodynamically trapped and which cells are outside of the trap, likely to be hydrodynamically washed out of the chip. Second, learning the difference between cells and traps might be enforced by explicitly learning to also segment each trap instance.

\subsection{Dataset Statistics}

Our full dataset is comprised of two distinct subsets, one for each of the trap geometries (\textit{regular} and \textit{L} traps). Details on the core features of both subsets are depicted in \cref{tab:dataset}. The first subset includes \textit{regular} trap types, and slight variations of this geometry, also referred to as type 1, whereas the second subset includes \textit{L}-shaped traps (type 2). In general, the first subset includes approximately four times the number of images and object instances (cells and traps) as the second subset.

\begin{table}[th!]
    \centering
    \caption{Core properties of our cells in microstructures dataset.}%
    \input{tables/dataset}
    \label{tab:dataset}
    \vspace{-0.5em}
\end{table}

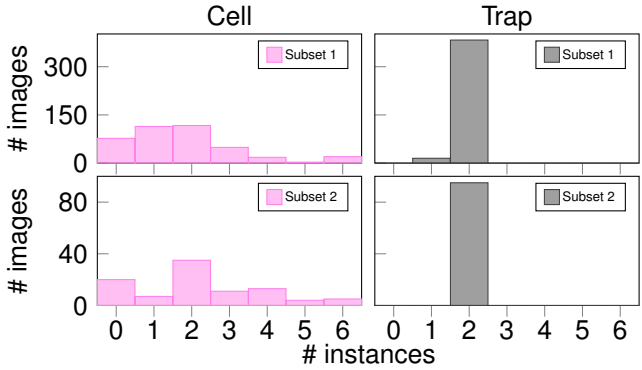
\begin{figure}[th!]
    \centering
    \input{artwork/instance_hist}
    \caption{Histogram showing the frequency of number of object instances in an image of our dataset. Left column visualizes the cell class (pink \colorindicator{cell}) and right column the trap class (grey \colorindicator{trap!50}).}
    \label{fig:instancecounthist}
    \vspace{-0.5em}
\end{figure}

We analyze the number of instances per class in each specimen image. \cref{fig:instancecounthist} shows the histogram of instances per image for both subsets and both semantic classes. The majority of images include two traps and at least a single cell. However, our dataset also includes specific edge cases, such as the case where only a single intact trap microstructure instance is present due to fabrication errors. In the most common setting, two cells and two trap instances are present. This corresponds to the setting of a trapped mother cell with a budding daughter cell that pinches off from the mother cell~\cite{Duina2014}. Images without any cells are also included. The maximum number of cells in a single image is six.

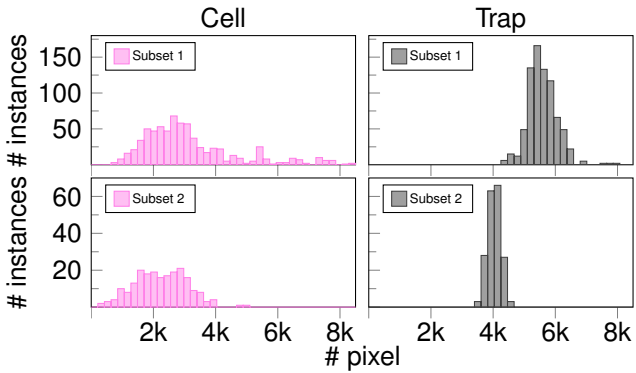
\begin{figure}[th!]
    \centering
    \input{artwork/object_size_hist}
    \caption{Histogram of object instance sizes in number of pixels. Left column visualizes the class cell (pink \colorindicator{cell}) and right column the class trap (grey \colorindicator{trap!50}).}
    \label{fig:objectsizehist}
    \vspace{-0.5em}
\end{figure}

Our dataset includes yeast cells of vastly different sizes. This is showcased in \cref{fig:objectsizehist}, where the cell size distribution approximately follows a normal distribution. The first subset, however, includes some outliers in the form of very large cells. The trap size histogram (\textit{cf.} \cref{fig:objectsizehist}) exhibits less variance than that of the cells (as is expected for microfabricated structures). The variation in trap appearance is included for increased model robustness and is due to a range of factors, ranging from fabrication tolerances, the position of the focal plane, to mechanical chip deformations (bending, warping, inclined mounting), amongst others. 

\begin{figure}[th!]
    \centering
    \input{artwork/cell_density}
    \caption{Density map of cell locations in our dataset. Pink (\colorindicator{cell}) areas indicate regions where many cells are located (H). White areas showcase regions where only a few or even no cells are located (L).}
    \label{fig:celldensity}
    \vspace{-0.5em}
\end{figure}
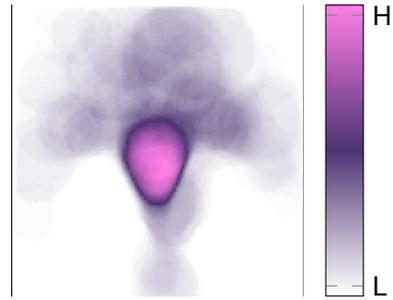

The distribution of cell positions is depicted as a density map in~\cref{fig:celldensity}. Yeast cells are mainly located inside a trap pair (\textit{cf.} \cref{fig:celldensity}). Additional cells are typically located above the trap pair. When budding, the daughter cell typically grows near the top of the microstructures. In some cases, daughter cells grow out of the bottom of the trap.

\subsection{Dataset Splits}

Our densely annotated microscopy images are split into three separate sets for training, validation, and testing. We initially split the dataset randomly. However, we subsequently manually curated the sets to ensure that all splits include a similar amount of variability in cell and trap configurations. Following the split fraction of the Cityscapes dataset ($\sim{}60\%$ training, $\sim{}10\%$ validation, and $\sim{}30\%$ test), we arrive at a split consisting of $296$ training, $49$ validation, $148$ test images with dense annotations. \cref{tab:datasetsplit} presents details of the dataset split.

\begin{table}[th!]
    \centering
    \caption{Training, validation, and test split of our dataset.}%
    \input{tables/dataset_split}
    \label{tab:datasetsplit}
    \vspace{-0.5em}
\end{table}

%% file: tables/dataset.tex
{
\begin{tabular*}{\columnwidth}{@{\extracolsep{\fill}}l@{}l@{}c@{}c@{}c@{}}
	\toprule
	 & Trap type & \# images & \# cells & \# traps \\
	\midrule
	Subset 1 & Type 1 (\textit{regular}) & 398 & 702 & 781 \\
	Subset 2 & Type 2 (\textit{L}) & \hphantom{2}95 & 212 & 190 \\
	\midrule
	Full dataset & Type 1 \& 2 & 493 & 914 & 971 \\
	\bottomrule
\end{tabular*}}

%% file: artwork/instance_hist.tex
\pgfplotsset{
/pgfplots/ybar legend/.style={
/pgfplots/legend image code/.code={\draw[#1, draw] (0cm,-0.1cm) rectangle ++ (0.2cm, 0.2cm);},
}
}
\begin{tikzpicture}[every node/.style={font=\fontsize{10}{10}\sffamily}]
    \node[anchor=center] at (0.418\columnwidth, -2.575) {\# instances\vphantom{p}};
    \begin{groupplot}[
        group style={
            group name=foo,
            group size=2 by 2,
            ylabels at=edge left,
            horizontal sep=5pt,
            vertical sep=5pt,
        },
        ylabel shift=-2.5pt,
        height=3.3cm,
        width=0.59\columnwidth,
        ybar,
        grid=none,
        xtick pos=bottom,
        ytick pos=left,
        ylabel=\# images,
        xtick={
            0, 1, 2, 3, 4, 5, 6
        },
        xticklabels={
            0, 1, 2, 3, 4, 5, 6
        },
        legend style={nodes={scale=0.5}, at={(0.95, 0.95)}},
        legend image code/.code={\draw[#1, draw] (0cm,-0.1cm) rectangle ++ (0.2cm, 0.2cm);},
        xmin=-0.5, xmax=6.5,
        x tick label style={yshift={0.075cm}},
        ]
        
        \nextgroupplot[title=\vphantom{p}Cell, title style={yshift=-8.25pt,}, ybar, bar width=14.2pt, ymax=402,ymin=0, ytick={0, 150, 300}, yticklabels={0, 150, 300}, grid=none, xticklabels={,,}]
        \addplot[cell, fill, fill opacity=0.5, ybar legend] coordinates { (0, 77) (1, 114) (2, 117) (3, 49) (4, 18) (5, 3) (6, 20) };
        \addlegendentry{Subset 1}

        \nextgroupplot[title=Trap, title style={yshift=-8.25pt,}, ybar, bar width=14.2pt, ymax=402,ymin=0, yticklabels={,,}, ytick={0, 150, 300}, grid=none, xticklabels={,,}]
         \addplot[trap, fill, fill opacity=0.5, ybar legend] coordinates { (0, 0) (1, 15) (2, 383) (3, 0) (4, 0) (5, 0) (6, 0) };
        \addlegendentry{Subset 1}
        
        \nextgroupplot[ybar, bar width=14.2pt, ymax=100, ymin=0, ytick={0, 40, 80}, yticklabels={0, 40, \hphantom{0}80}, grid=none]
        \addplot[cell, fill, fill opacity=0.5, ybar legend] coordinates { (0, 20) (1, 7) (2, 35) (3, 11) (4, 13) (5, 4) (6, 5) };
        \addlegendentry{Subset 2}

        \nextgroupplot[ybar, bar width=14.2pt, ymax=100, ymin=0, yticklabels={,,}, ytick={0, 40, 80}, grid=none]
        \addplot[trap, fill, fill opacity=0.5, ybar legend] coordinates { (0, 0) (1, 0) (2, 95) (3, 0) (4, 0) (5, 0) (6, 0) };
        \addlegendentry{Subset 2}
        
    \end{groupplot}
\end{tikzpicture}%

%% file: artwork/object_size_hist.tex
\begin{tikzpicture}[every node/.style={font=\fontsize{10}{10}\sffamily}]
    \node[anchor=center] at (0.418\columnwidth, -2.575) {\# pixel};
    \begin{groupplot}[
        group style={
            group name=foo,
            group size=2 by 2,
            ylabels at=edge left,
            horizontal sep=5pt,
            vertical sep=5pt,
        },
        ylabel shift=-2.5pt,
        height=3.3cm,
        width=0.59\columnwidth,
        ybar,
        major grid style={line width=.2pt,draw=gray!50},
        minor y tick num=1,
        xtick pos=bottom,
        ytick pos=left,
        xmin=0,
        xmax=8500,
        ylabel=\# instances,
        xtick={
            0, 2000, 4000, 6000, 8000
        },
        xticklabels={
            , 2k, 4k, 6k, 8k
        },
        legend style={nodes={scale=0.5}, at={(0.395, 0.95)}},
        legend image code/.code={\draw[#1, draw] (0cm,-0.1cm) rectangle ++ (0.2cm, 0.2cm);}
        xmin=0, xmax=8500,
        x tick label style={yshift={0.075cm}},
        ]
        \nextgroupplot[title=\vphantom{p}Cell, title style={yshift=-8.25pt,}, ytick={0, 50, 100, 150, 200}, yticklabels={, 50, 100, 150,}, ymin=0, ymax=180, xticklabels={,,}]
        \addplot +[cell, hist={bins=40}, fill opacity=0.5] table [y index=0] {cell_sizes_r.csv};
        \addlegendentry{Subset 1}
        \nextgroupplot[title=Trap, title style={yshift=-8.25pt,}, ytick={0, 50, 100, 150, 200}, yticklabels={,,}, xtick={0, 2000, 4000, 6000, 8000}, ymin=0, ymax=180, xticklabels={,,}]
        \addplot +[trap, hist={bins=40}, fill opacity=0.5] table [y index=0] {trap_sizes_r.csv};
        \addlegendentry{Subset 1}
        
        \nextgroupplot[ytick={0, 20, 40, 60}, yticklabels={, 20, 40, \hphantom{0}60}, ymin=0, ymax=70]
        \addplot +[cell, hist={bins=40}, fill opacity=0.5] table [y index=0] {cell_sizes_l.csv};
        \addlegendentry{Subset 2}
        \nextgroupplot[ytick={0, 20, 40, 60}, yticklabels={,,}, ymin=0, ymax=70]
        \addplot +[trap, hist={bins=40}, fill opacity=0.5] table [y index=0] {trap_sizes_l.csv};
        \addlegendentry{Subset 2}
        
    \end{groupplot}
\end{tikzpicture}%

%% file: artwork/cell_density.tex
\begin{tikzpicture}[every node/.style={font=\fontsize{10}{10}\sffamily}]

\definecolor{darkgray176}{RGB}{176,176,176}

\begin{axis}[
colorbar,
colorbar style={
ylabel={},
ytick={12, 337},
yticklabels={L, H}},
colormap={mymap}{[1pt]
  rgb(0pt)=(1,1,1);
  rgb(1pt)=(0.3,0.2,0.45);
  rgb(2pt)=(1,0.5,0.90980392)
},
ticks=none,
xtick=\empty, 
ytick=\empty,
height=5.45cm,
point meta max=349,
point meta min=0,
tick align=outside,
tick pos=left,
width=5.45cm,
x grid style={black},
xmin=-0.5, xmax=255.5,
xtick style={color=black},
y dir=reverse,
y grid style={black},
ymin=-0.5, ymax=255.5,
ytick style={color=black},
]
\addplot graphics [includegraphics cmd=\pgfimage,xmin=0, xmax=255.5, ymin=255.5, ymax=-0.5] {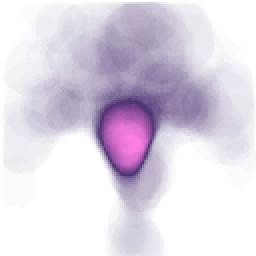};
\end{axis}
\end{tikzpicture}%

%% file: tables/dataset_split.tex
{
\begin{tabular*}{\columnwidth}{@{\extracolsep{\fill}}l@{}c@{}c@{}c@{}c@{}}
	\toprule
	Split & \# images & \# cells & \# traps & Trap type images 1 \textit{vs.} 2 \\
	\midrule
	Training & 296 & 536 & 582 & 244/52 \\
	Validation & \hphantom{2}49 & 108 & \hphantom{2}98 & \hphantom{2}33/16 \\
	Test & 148 & 270 & 291 & 121/27 \\
	\bottomrule
\end{tabular*}}

%% file: content/performance_evaluation.tex
\section{Performance Evaluation} \label{sec:performanceevaluation}

We propose to utilize both the cell class intersection-over-union ($\operatorname{IoU}$) and the panoptic quality ($\operatorname{PQ}$) metrics~\cite{Kirillov2019} to evaluate instance segmentation algorithms on our dataset. The cell class $\operatorname{IoU}$ (Jaccard index) is defined as:

\begin{equation}
    \operatorname{IoU}\left(p_{\rm c}, g_{\rm c}\right) = \frac{\left|p_{\rm c}\cap g_{\rm c}\right|}{\left|p_{\rm c}\cup g_{\rm c}\right|},
\end{equation}
where $p_{\rm c}$ is the segment for the cell class and $g_{\rm c}$ indicates the cell class ground truth. This metric evaluates the semantic performance of the cell segmentation. While we are presenting an instance segmentation dataset, we propose to validate the performance of instance segmentation approaches on our dataset with the cell class $\operatorname{IoU}$ to ensure a comparison between previous work which utilizes this metric~\cite{Bakker2018, Prangemeier2020a, Prangemeier2020b, Prangemeier2022}. The cell class $\operatorname{IoU}$ is biased towards large objects and does not capture the recognition and segmentation of individual object instances. However, when an application does not require single-cell segmentations but the semantic segmentation of all cells, this metric is an insightful measure of the semantic segmentation performance.

To measure the instance segmentation performance on our yeast cells in microstructures dataset, we propose to utilize the $\operatorname{PQ}$ metric. In panoptic segmentation, object classes are categorized into two different class types. First, ``stuff classes'' include uncountable objects/regions such as sidewalks or grass. Second, ``things classes'' include countable objects classes, such as cars, people, or bicycles. For stuff classes, semantic segmentation is performed, whereas instance segmentation is performed for things classes. Our instance segmentation dataset can be seen as an edge-case of panoptic segmentation. Cell and trap classes can be viewed as things classes, whereas the background builds the only stuff class. Thus, we can evaluate an instance segmentation prediction for our dataset with the $\operatorname{PQ}$.

The $\operatorname{PQ}$ is computed individually for each semantic class (background, cell, \& trap) and averaged over all classes. Before computing the $\operatorname{PQ}$ for a class, the predicted and labeled instances are matched. This matching results in three sets: true positive ($\rm TP$), false positive ($\rm FP$), and false negative ($\rm FN$) matches. Please refer to Kirillov \textit{et al.}~\cite{Kirillov2019} for details on the matching approach. Based on these sets the $\operatorname{PQ}$ is computed for each semantic class as:

\begin{equation}
    \operatorname{PQ}=\underbrace{\frac{\sum_{(p,g)\in {\rm TP}}\operatorname{IoU}(p, g)}{\vphantom{\frac{1}{2}}\left|{\rm TP}\right|}}_{\operatorname{SQ}}\, \underbrace{\frac{\left|{\rm TP}\right|}{\left|{\rm TP}\right| + \frac{1}{2}\left|{\rm FP}\right| + \frac{1}{2}\left|{\rm FN}\right|}}_{\operatorname{RQ}},
\end{equation}
here $\frac{1}{\left|{\rm TP}\right|}\sum_{(p,g)\in {\rm TP}}\operatorname{IoU}(p, g)$ computes the mean $\operatorname{IoU}$ of all matched predicted $p$ and ground truth $g$ segments. The $\operatorname{PQ}$ is a measure of both the (instance-wise) segmentation quality ($\operatorname{SQ}$) and the recognition quality ($\operatorname{RQ}$) of a panoptic segmentation prediction. Additionally, the $\operatorname{PQ}$ weights each object instance importance independent of their size.

By measuring both the cell class $\operatorname{IoU}$ and the $\operatorname{PQ}$, we can evaluate the performance of segmentation algorithms on our instance segmentation dataset. For applications requiring single-cell segmentations and the positioning and segmentation of microstructures, the $\operatorname{PQ}$ is the superior metric. The cell class $\operatorname{IoU}$ is to be preferred for applications requiring only semantic information of cells. We offer code for computing both the cell class $\operatorname{IoU}$ and the panoptic quality for a standardized comparison of new approaches.

%% file: content/conclusion_outlook.tex
\section{Conclusion and Outlook} \label{sec:conclusionoutlook}

In this paper, we presented a new dataset for segmenting yeast cells in microstructures, a widespread scenario for a key model organism in biological research and development. We provide both pixel-wise instance segmentation labels and a standardized performance evaluation strategy. The aim of this joint approach is to facilitate progress in the field of trapped yeast analysis and to provide a basis for a fair comparison between instance segmentation methods.

Beyond the scenario presented here, some biomedical applications require temporal cell segmentations. To aid the development of unified cell segmentation and tracking algorithms, future work may consider extending our dataset with video instance segmentation labels~\cite{Yang2019}.